%% file: root.tex
\documentclass[conference]{IEEEtran}
\usepackage{times}

\usepackage[numbers]{natbib}
\usepackage{multicol}
\usepackage[bookmarks=true, hidelinks]{hyperref}

\usepackage{graphics} 
\usepackage{graphicx}
\usepackage[table]{xcolor}
\usepackage{url}
\usepackage{color}

\usepackage{amsmath}
\usepackage{mathtools}
\usepackage{amsfonts}
\usepackage{algorithm2e, setspace}

\usepackage{multirow}
\usepackage{soul}
\usepackage{adjustbox}

\graphicspath{{./}{./pics/}{../pics}}
\DeclareGraphicsExtensions{.jpg,.JPG,.png,.PNG,.pdf,.PDF,.eps,.EPS}

\newcommand{\prbpf}[0]{PoseRBPF}

\begin{document}

\newcommand{\todo}[1]{\textcolor{red}{\textbf{#1}}}

\title{\prbpf: A Rao-Blackwellized Particle Filter for  6D Object  
Pose Tracking}

\author{Author Names Omitted for Anonymous Review. Paper-ID 152}


%
\author{\authorblockN{
Xinke Deng\authorrefmark{1}\authorrefmark{2},
Arsalan Mousavian\authorrefmark{1},
Yu Xiang\authorrefmark{1}, 
Fei Xia\authorrefmark{1}\authorrefmark{3},
Timothy Bretl\authorrefmark{2},
Dieter Fox\authorrefmark{1}\authorrefmark{4}}
\authorblockA{\authorrefmark{1}NVIDIA \hspace*{5ex} \authorrefmark{2}University of Illinois at Urbana-Champaign}
\authorblockA{\authorrefmark{3}Stanford University\hspace*{5ex} \authorrefmark{4}University of Washington}
}

\maketitle

\begin{abstract}
 
Tracking 6D poses of objects from videos provides rich information to a robot in performing different tasks such as manipulation and navigation. In this work, we formulate the 6D object pose tracking problem in the Rao-Blackwellized particle filtering framework, where the 3D rotation and the 3D translation of an object are decoupled. This factorization allows our approach, called \prbpf\, to efficiently estimate the 3D translation of an object along with the full distribution over the 3D rotation. This is achieved by discretizing the rotation space in a fine-grained manner, and training an auto-encoder network to construct a codebook of feature embeddings for the discretized rotations. As a result, \prbpf\ can track objects with arbitrary symmetries while still maintaining adequate posterior distributions.  Our approach achieves state-of-the-art results on two 6D pose estimation benchmarks. A video showing the experiments can be found at {\color{blue}\href{https://youtu.be/lE5gjzRKWuA}{https://youtu.be/lE5gjzRKWuA}}. 

\end{abstract}

\IEEEpeerreviewmaketitle

\input{0_intro}

\input{1_related_work}

\input{2_methodology}

\input{3_experiments}

\section{CONCLUSION} 
\label{sec:conclusion}
In this work, we introduced \prbpf, a Rao-Blackwellized particle filter for tracking 6D object poses. Each particle samples 3D translation and estimates the distribution over 3D rotations conditioned on the image bonding box corresponding to the sampled translation. \prbpf\ compares each bounding box embedding to learned viewpoint embeddings so as to efficiently update distributions over time.  We demonstrated that the tracked distributions capture both the uncertainties from the symmetry of objects and the uncertainty from object pose. Experiments on two benchmark datasets with house hold objects and  symmetric texture less industrial objects show the superior performance of \prbpf.




\bibliographystyle{plainnat}
\bibliography{references}

\end{document}

%% file: 0_intro.tex
\section{INTRODUCTION}

Estimating the 6D pose of objects from camera images, i.e., 3D rotation and 3D translation of an object with respect to the camera, is an important problem in robotic applications. For instance, in robotic manipulation, 6D pose estimation of objects provides critical information to the robot for planning and executing grasps. In robotic navigation tasks, localizing objects in 3D provides useful information for planing and obstacle avoidance. Due to its significance, various efforts have been devoted to tackling the 6D pose estimation problem from both the robotics community \cite{collet2011moped,cao2016real,xiang2017posecnn,tremblay2018corl:dope} and the computer vision community \cite{rothganger20063d,liebelt2008independent,hinterstoisser2012model}.

Traditionally, the 6D pose of an object is estimated using local-feature or template matching techniques, where features extracted from an image are matched against features or viewpoint templates generated for the 3D model of the object.  The 6D object pose can then be recovered using 2D-3D correspondences of these local features or by selecting the best matching viewpoint onto the object~\cite{collet2011moped,hinterstoisser2012gradient,hinterstoisser2012model}. More recently, machine learning techniques have been employed to detect key points or learn better image features for matching \cite{brachmann2014learning,krull2015learning}. Thanks to advances in deep learning, convolutional neural networks have recently been shown to significantly boost the estimation accuracy and robustness~\cite{kehl2017ssd,zeng2017multi,pavlakos20176,tekin2018real,xiang2017posecnn}, 

So far, the focus of image-based 6D pose estimation has been on the \emph{accuracy of single image estimates}; most techniques ignore temporal information and provide only a single hypothesis for an object pose. In robotics, however, temporal data and information about the \emph{uncertainty} of estimates can also be very important for tasks such as grasp planning or active sensing. Temporal tracking in video data can improve pose estimation~\cite{oka2009tracking,choi20123d,krull20146,crivellaro2015novel}. In the context of point-cloud based pose estimation, Kalman filtering has also been used to track 6D poses, where Bingham distributions have been shown to be well suited for orientation estimation~\cite{Sri17Bin}. However, unimodal estimates are not sufficient to adequately represent the complex uncertainties arising from occlusions and possible object symmetries. 

\begin{figure}
    \centering
    \includegraphics[width=0.45\textwidth]{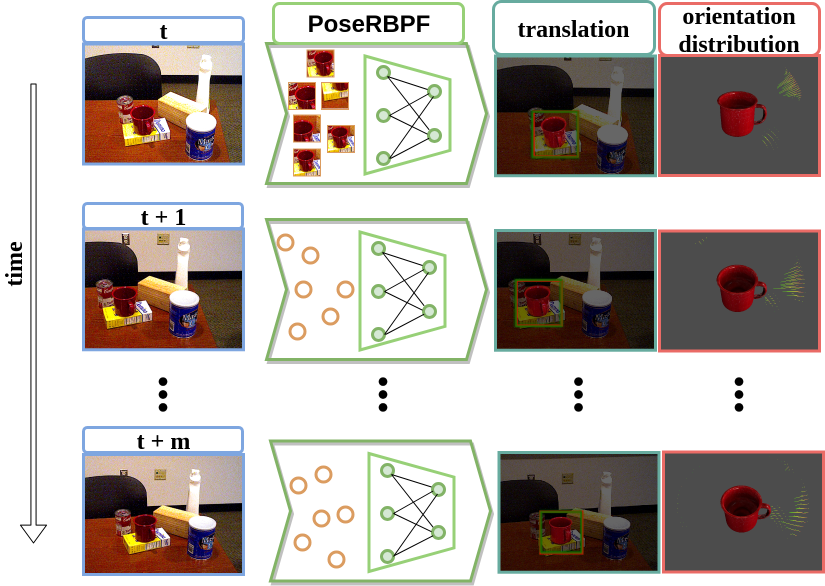}
    \caption{Overview of our PoseRBPF framework for 6D object pose tracking. Our method leverages a Rao-Blackwellized particle filter and an auto-encoder network to estimate the 3D translation and a full distribution of the 3D rotation of a target object from a video sequence.}
    \label{fig:intro}
    \vspace{-4mm}
\end{figure}

In this work, we introduce a particle filter-based approach to estimate full posteriors over 6D object poses.  Our approach, called \prbpf, factorizes the posterior into the 3D translation and the 3D rotation of the object, and uses a Rao-Blackwellized particle filter that samples object poses and estimates discretized distributions over rotations for each particle. To achieve accurate estimates, the 3D rotation is discretized at 5 degree resolution, resulting in a distribution over $72\times37\times72= 191,808$ bins for each particle (elevation ranges only from -90 to 90 degree). To achieve real time performance, we pre-compute a codebook over  embeddings for all discretized rotations, where embeddings come from an auto-encoder network trained to encode the visual appearance of an object from arbitrary viewpoints at a certain scale (inspired by \cite{sundermeyer2018implicit}). For each particle, \prbpf\ first uses the 3D translation to determine the center and size of the object bounding box in the image, then determines the embedding for that bounding box, and finally updates the rotation distribution by comparing the embedding value with the pre-computed entries in the codebook using cosine distance. The weight of each particle is given by the normalization factor of the rotation distribution. Motion updates are performed efficiently by sampling from a motion model over poses and a convolution over the rotations. Fig.~\ref{fig:intro} illustrates our PoseRBPF framework for 6D object pose tracking. Experiments on the YCB-Video dataset \cite{xiang2017posecnn} and the T-Less dataset \cite{hodan2017t} show that PoseRBPFs are able to represent uncertainties arising from various types of object symmetries and can provide more accurate 6D pose estimation.

Our work makes the following main contributions:
\begin{itemize}
    \item We introduce a novel 6D object pose estimation framework that combines Rao-Blackwellized particle filtering with a learned auto-encoder network in an efficient and principled way.
    \item Our framework is able to track full distributions over 6D object poses.  It can also do so for objects with arbitrary kinds of symmetries, without the need for any manual symmetry labeling. 
\end{itemize}
The rest of the paper is organized as follows. After discussing the related work, we present our Rao-Blackwellized particle filtering framework for 6D object pose tracking, followed by experimental evaluations and a conclusion.



%% file: 1_related_work.tex
\section{RELATED WORK}

Our work is closely related to recent advances in 6D object pose estimation using deep neural networks. The current trend is to augment state-of-the-art 2D object detection networks with the ability to estimate 6D object pose. For instance, \cite{kehl2017ssd} extend the SSD detection network \cite{liu2016ssd} to 6D pose estimation by adding viewpoint classification to the network. \cite{tekin2018real} utilize the YOLO architecture \cite{redmon2016you} to detect 3D bounding box corners of objects in the images, and then recover the 6D pose by solving the PnP problem. PoseCNN \cite{xiang2017posecnn} designs an end-to-end network for 6D object pose estimation based on the VGG architecture \cite{simonyan2014very}. Although these methods significantly improve the 6D pose estimation accuracy over the traditional methods \cite{hinterstoisser2012model,brachmann2014learning,krull2015learning}, they still face difficulty in dealing with symmetric objects, where most methods manually specify the symmetry axis for each such object. In contrast, \cite{sundermeyer2018implicit} introduce an implicit way of representing 3D rotations by training an auto-encoder for image reconstruction, which does not need to pre-define the symmetry axes for symmetric objects. We leverage this implicit 3D rotation representation in our work, and show how to combine it with particle filtering for 6D object pose tracking.

The particle filtering framework has been widely applied to different tracking applications in the literature \cite{nummiaro2002object,shan2004real,khan2005mcmc,sarkka2007rao}, thanks to its flexibility in incorporating different observation models and motion priors. Meanwhile, it offers a rigorous probabilistic formulation to estimate uncertainty in the tracking results. Different approaches have also been proposed to track the poses of objects using particle filters \cite{azad20116,choi2012robust,pauwels2013real,xiang2014monocular,li2015real}. However, in order to achieve good tracking performance, a particle filter requires a strong observation model. Also, the tracking frame rate is limited by the particle sampling efficiency. In this work, we factorize the 6D object pose tracking problem and deploy Rao-Blackwellized particle filters~\cite{doucet2000rao}, which have been shown to scale to complex estimation problems such as SLAM~\cite{PRbook,Mon02Fas} and multi-model target tracking~\cite{Kwo04Map,sarkka2007rao}.  We also employ a deep neural network as an observation model that provides robust estimates for object orientations even under occlusions and symmetries. Our design allows us to evaluate all possible orientations in parallel using an efficient GPU implementation. As a result, our method can track the distribution of the 6D pose of an object at 20fps.

%% file: 2_methodology.tex
\section{6D Object Pose Tracking with PoseRBPF}

The goal of 6D object pose tracking of an object is to estimate the 3D rotation $\mathbf{R}$ and the 3D translation $\mathbf{T}$ of the object for every frame in an image stream. In this section, we first formulate the 6D object tracking problem in a particle filtering framework, and then describe how to utilize a deep neural network to compute the likelihoods of the particles and to achieve an efficient sampling strategy for tracking. 

\begin{figure}
    \centering
    \includegraphics[width=0.5\textwidth]{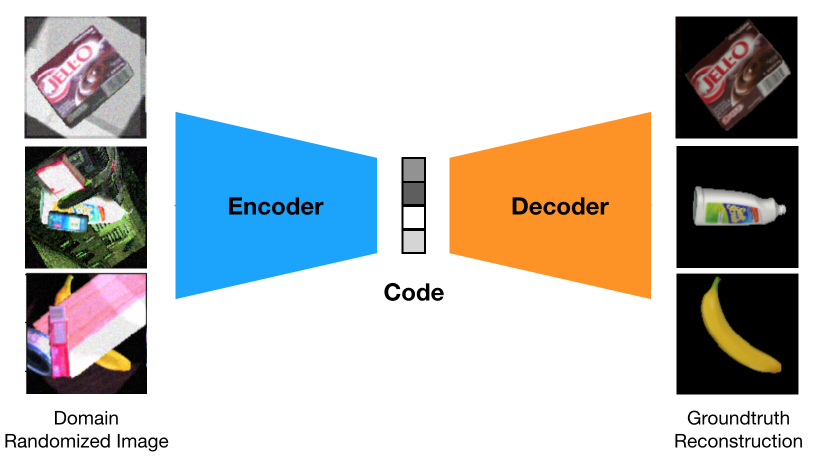}
    \caption{Illustration of the inputs and outputs of the auto-encoder. Images with different lighting, background and occlusion are feed into the network to reconstruct synthetic images of the objects from the same 6D poses. The encoder generates a feature embedding (code) of the input image.}
    \label{fig:autoencoder}
    \vspace{-3mm}
\end{figure}
 
\subsection{Rao-Blackwellized Particle Filter Formulation}
\begin{figure*}
    \centering
    \includegraphics[width=0.8\textwidth]{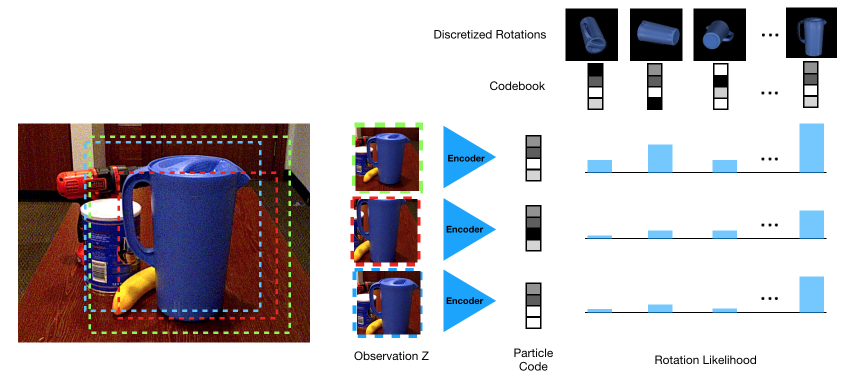}
    \caption{Illustration of the computation for the conditional rotation likelihood by codebook matching. Left) Each particle crops the image based on its translation hypothesis. The RoI for each particle is resized and the corresponding code is computed using the encoder.  Right) The rotation distribution $P(\mathbf{R}|\mathbf{Z}, \mathbf{T})$ is computed from the  distance between the code for each hypothesis and those in the codebook.}
    \label{fig:observation_likelihood} 
    \vspace{-4mm}
\end{figure*}

At time step $k$, given observations $\mathbf{Z}_{1:k}$ up to time $k$, our primary goal is to estimate the posterior distribution of the 6D pose of an object $P(\mathbf{R}_k, \mathbf{T}_k|\mathbf{Z}_{1:k})$, where $\mathbf{R}_k$ and $\mathbf{T}_k$ denote the 3D rotation and 3D translation of the object at time $k$, respectively.  Using a vanilla particle filter to sample over this 6D space is not feasible, especially when there is large uncertainty over the orientation of the object. Such uncertainties occur frequently when objects are heavily occluded or have symmetries that result in multiple orientation hypotheses. We thus propose to factorize the 6D pose estimation problem into 3D rotation estimation and 3D translation estimation. This idea is based on the observation that the 3D translation can be estimated from the location and the size of the object in the image. The translation estimation provides the center and scale of the object in the image, based on which the 3D rotation can be estimated from the appearance of the object inside the bounding box. Specifically, we decompose the posterior into:
\begin{equation}
    P(\mathbf{R}_k, \mathbf{T}_k|\mathbf{Z}_{1:k}) =  P(\mathbf{T}_k|\mathbf{Z}_{1:k}) P(\mathbf{R}_k | \mathbf{T}_k, \mathbf{Z}_{1:k}),
\end{equation}
where $P(\mathbf{T}_k|\mathbf{Z}_{1:k})$ encodes the location and scale of the object, and $P(\mathbf{R}_k | \mathbf{T}_k, \mathbf{Z}_{1:k})$ models the rotation distribution conditioned on the translation and the images.

This factorization directly leads to an efficient sampling scheme for a Rao-Blackwellized particle filter~\cite{doucet2000rao,PRbook}, where the posterior at time $k$ is approximated by a set of $N$ weighted samples $\mathcal{X}_{k} = \{ \mathbf{T}_{k}^i,  P( \mathbf{R}_{k}|{\mathbf{T}_{k}^i, \mathbf{Z}_{1:k}}), w_{k}^i \}_{i=1}^N$. Here, $\mathbf{T}_{k}^i$ denotes the translation of the $i$th particle, $P( \mathbf{R}_{k}|{\mathbf{T}_{k}^i, \mathbf{Z}_{1:k}})$ denotes the discrete \emph{distribution} of the particle over the object orientation conditioned on the translation and the images, and $w_{k}^i$ is the importance weight. To achieve accurate pose estimation, the 3D object orientation consisting of azimuth, elevation, and in-plane rotation is discretized into bins of size 5 degree, resulting in a distribution over $72\times37\times72= 191,808$ bins for each particle (elevation ranges only from -90 to 90 degrees). At every time step $k$, the particles are propagated through a motion model to generate a new set of particles $\mathcal{X}_{k+1}$, from which we can estimate the 6D pose distribution.

\subsection{Observation Likelihoods} \label{sec:observation_model}

The observation likelihoods of the two posteriors $P(\mathbf{Z}_k | \mathbf{T}_k)$ and $P(\mathbf{Z}_k| {\mathbf{T}_k}, \mathbf{R}_k)$ measure the compatibility of the observation $\mathbf{Z}_k$ with the object pose at the 3D rotation $\mathbf{R}_k$ and the 3D translation $\mathbf{T}_k$. According to the Bayes Rule
\begin{equation}
    P(\mathbf{Z}_k | \mathbf{T}_k, \mathbf{R}_k) \propto P(\mathbf{R}_k|\mathbf{T}_k, \mathbf{Z}_k)P(\mathbf{Z}_k|\mathbf{T}_k), \label{eq:ob_likelihood}
\end{equation} 
it is sufficient to estimate the likelihood $P(\mathbf{Z}_k | \mathbf{T}_k, \mathbf{R}_k)$ by computing $P(\mathbf{R}_k|\mathbf{T}_k, \mathbf{Z}_k)$ and $P(\mathbf{Z}_k|\mathbf{T}_k)$. Intuitively, a 6D object pose estimation method, such as \cite{kehl2017ssd,tekin2018real,xiang2017posecnn}, can be employed to estimate the observation likelihoods. However, these methods only provide a single estimation of the 6D pose instead of estimating a probability distribution, i.e., there is no uncertainty in their estimation. Also, these methods are computationally expensive if we would like to evaluate a large number of samples in the particle filtering.

Ideally, if we can synthetically generate an image of the object with the pose $(\mathbf{R}_k, \mathbf{T}_k)$ into the same scene as the observation $\mathbf{Z}_k$, we can compare the synthetic image with the input image $\mathbf{Z}_k$ to measure the likelihoods. However, this is not feasible since it is very difficult to synthesize the same lighting, background or even occlusions between objects as in the input video frame. In contrast, it is straightforward to render a synthetic image of the object using constant lighting, blank background and no occlusion, given the 3D model of the object. Therefore, inspired by \cite{sundermeyer2018implicit}, we apply an auto-encoder to transform the observation $\mathbf{Z}_k$ into the same domain as the synthetic rendering of the object. Then we can compare image features in the synthetic domain to measure the likelihoods of 6D poses efficiently.

\subsubsection{Auto-encoder}

An auto-encoder is trained to map an image $\mathbf{Z}$ of the target object with pose $(\mathbf{R}, \mathbf{T})$ to a synthetic image $\mathbf{Z}'$ of the object rendered from the same pose, where the synthetic image $\mathbf{Z}'$ is rendered using constant lighting, and there is no background and occlusion in the synthetic image. In this way, the auto-encoder is forced to map images with different lighting, background and occlusion to the common synthetic domain. Fig. \ref{fig:autoencoder} illustrates the input and output of the auto-encoder during training. In addition, the auto-encoder learns a feature embedding $f(\mathbf{Z})$ of the input image.

Instead of training the auto-encoder to reconstruct images with arbitrary 6D poses, which makes the training challenging, we fix the 3D translation to a canonical one $\mathbf{T}_0 = (0, 0, z)^T$, where $z$ is a pre-defined constant distance. The canonical translation indicates that the target object is in front of the camera with distance $z$. The 3D rotation $\mathbf{R}$ is uniformly sampled during training. After training, for each discretized 3D rotation $\mathbf{R}^i$, a feature embedding $f(\mathbf{Z}(\mathbf{R}^i, \mathbf{T}_0))$ is computed using the encoder, where $\mathbf{Z}(\mathbf{R}^i, \mathbf{T}_0)$ denotes a rendered image of the target object from pose $(\mathbf{R}^i, \mathbf{T}_0)$. We consider the set of all the feature embeddings of the discretized 3D rotations to be the codebook of the target, and we show how to compute the likelihoods using the codebook next.

\begin{figure}
\centering
    \includegraphics[width=0.45\textwidth]{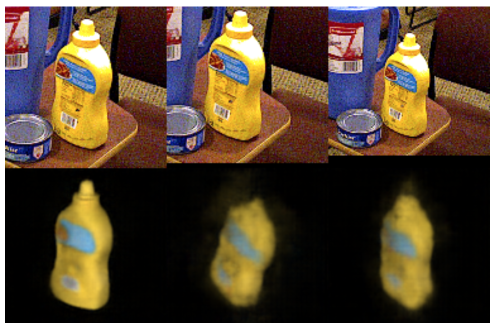}
    \caption{Visualization of reconstruction of the RoIs from auto-encoder. Left is the groundtruth RoI. The other two column show the reconstruction with shifting and scale change. As it is shown the reconstruction quality degrades with deviations from groundtruth RoI. This property makes auto-encoder a suitable choice for computing the observation likelihoods.}
    \label{fig:shifting_effect}
    \vspace{-4mm}
\end{figure} 

\begin{figure*}
    \centering
    \includegraphics[width=0.85\textwidth]{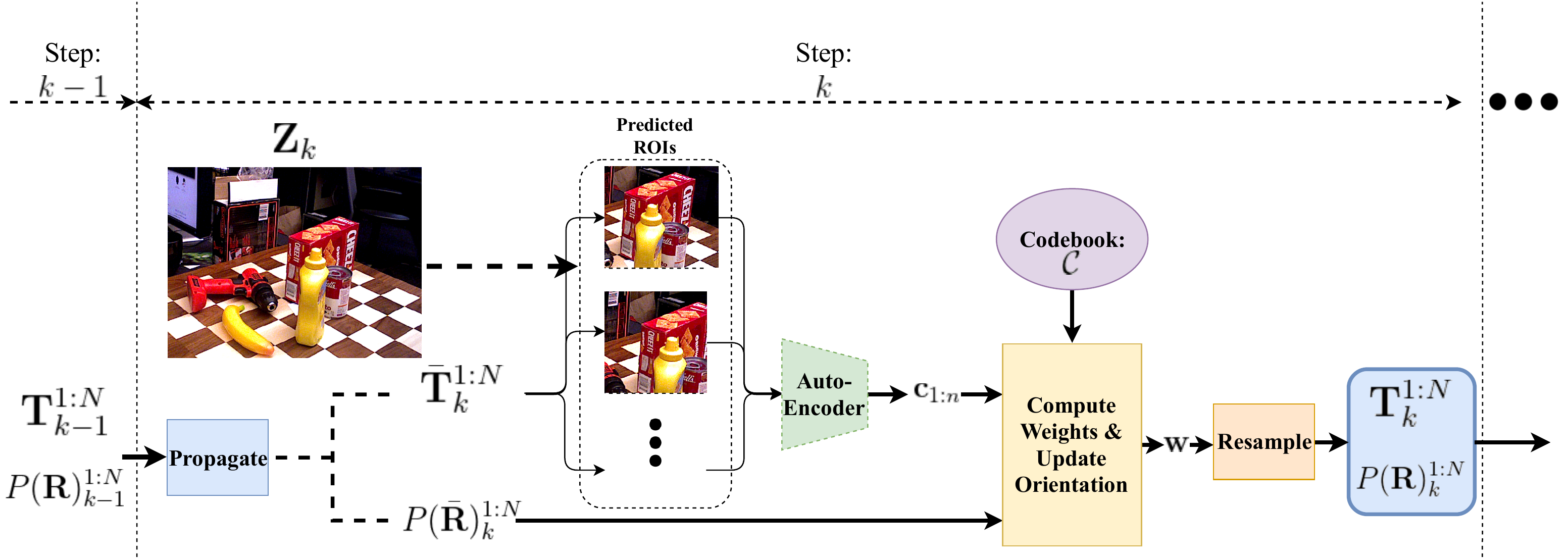}
    \caption{We propose PoseRBPF, a Rao-Blackwellized particle filter for 6D object pose tracking. For each particle, the \emph{orientation distribution} is estimated conditioned on translation estimation, while the translation estimation is evaluated with the corresponding RoIs.}
    \label{fig:framework}
    \vspace{-2mm}
\end{figure*}  

\subsubsection{Codebook Matching}

Given a 3D translation hypothesis $\mathbf{T}_k$, we can crop a Region of Interest (RoI) from the image $\mathbf{Z}_k$, and then feed the RoI into the encoder to compute a feature embedding of the RoI. Specifically, the 3D translation $\mathbf{T}_k = (x_k, y_k, z_k)^T$ is projected to the image to find the center $(u_k, v_k)$ of the RoI :
\begin{equation} \label{eq:projection}
\begin{bmatrix}
u_k \\[0.5em] v_k
\end{bmatrix} = \begin{bmatrix}
f_x \frac{x_k}{z_k} + p_x \\[0.5em]
f_y \frac{y_k}{z_k} + p_y
\end{bmatrix},
\end{equation}
where $f_x$ and $f_y$ indicate the focal lengths of the camera, and $(p_x, p_y)^T$ is the principal point. The size of the RoI is determined by $\frac{z_k}{z} s$, where $z$ and $s$ are the canonical distance and the RoI size in training the auto-encoder, respectively. Note that each RoI is a square region in our case, which makes the RoI independent from the rotation of the object.


The RoI is feed into the encoder to compute the feature embedding $f(\mathbf{Z}_k(\mathbf{T}_k))$. Finally, we compute the cosine distance, which is also referred as a similarity score, between the feature embedding of the RoI and a code in the codebook to measure the rotation likelihood:
\begin{equation} \label{eq:similarity}
    P(\mathbf{R}_c^j|\mathbf{Z}_k, \mathbf{T}_k) \propto \phi\Big(\frac{f(\mathbf{Z}_k(\mathbf{T}_k)) \cdot f(\mathbf{Z}(\mathbf{R}_c^j, \mathbf{T}_0))}{\|f(\mathbf{Z}_k(\mathbf{T}_k))\| \cdot \|f(\mathbf{Z}(\mathbf{R}_c^j, \mathbf{T}_0))\|}\Big),
\end{equation}
where $\mathbf{R}_c^j$ is one of the discretized rotations in the codebook, and $\phi(\cdot)$ is a Gaussian probability density function centered at the maximum cosine distance among all the codes in the codebook for all the particles. In this way, we can obtain a probabilistic likelihood distribution of all the rotations in the codebook given a translation. Fig.~\ref{fig:observation_likelihood} illustrates the computation of the rotation likelihoods by the cookbook matching.

Since the auto-encoder is trained with the object being at the center of the image and at a certain scale, i.e., with the canonical translation $\mathbf{T}_0$, any change in scale or deviation of the object from the center of the image results in poor reconstructions (see Fig.~\ref{fig:shifting_effect}). Particles with incorrect translations would generate RoIs where the object is not in the center of the RoI or with the wrong scale. Then we can check the reconstruction quality of the RoI to measure the likelihood of the translation hypothesis. We utilize this property to compute the translation likelihood $P(\mathbf{Z}_k | \mathbf{T}_k)$. Intuitively, if the translation $\mathbf{T}_k$ is correct, the similarity scores in Eq.~\eqref{eq:similarity} for rotation $\mathbf{R}^i$ that is close to the ground truth rotation would be high.
Specifically, $P(\mathbf{Z}_k | \mathbf{T}_k)$ is computed as the sum of the probability density $P(\mathbf{R}_c^j|\mathbf{T}_k, \mathbf{Z}_k)$ for all the discrete rotations.

\subsection{Motion Priors}\label{sec:motion_prior}

Motion prior is used to propagate the distribution of the poses from the previous time step $k-1$ to the current time step $k$. We use a constant velocity model to propagate the probability distribution of the 3D translation:
\begin{equation}
    P(\mathbf{T}_k|\mathbf{T}_{k-1}, \mathbf{T}_{k-2}) = \mathcal{N}\left(\mathbf{T}_{k-1} + \alpha (\mathbf{T}_{k-1} - \mathbf{T}_{k-2}), \mathbf{\Sigma_T} \right),
\end{equation}
where $\mathcal{N}(\mathbf{\mu}, \mathbf{\Sigma})$ denotes the multivariate normal distribution with mean $\mathbf{\mu}$ and covariance matrix $\mathbf{\Sigma}$, and $\alpha$ is a hyper-parameter of the constant velocity model.
The rotation prior is defined as a normal distribution with mean $\mathbf{R}_{k-1}$ and fixed covariance $\mathbf{\Sigma_R}$:
\begin{equation}
    P(\mathbf{R}_k|\mathbf{R}_{k-1}) = \mathcal{N}\left(\mathbf{R}_{k-1}, \mathbf{\Sigma_R} \right),
\end{equation}
where we represent the rotation $\mathbf{R}$ using Euler angles. Then the rotation prior can be implemented by a convolution on the previous rotation distribution with a 3D Gaussian kernel.


\subsection{6D Object Pose Tracking Framework} \label{sec:tracking}

The tracking process can be initialized from any 2D object detector that outputs a 2D bounding box of the target object. Given the first frame $\mathbf{Z}_1$, we backproject the center of the 2D bounding box to compute the $(x, y)$ components of the 3D translation and sample different $z$s uniformly to generate a set of translation hypotheses. The translation $\mathbf{T}_1$ with the highest likelihood $P(\mathbf{Z}_1 | \mathbf{T})$ is used as the initial hypothesis and $P(\mathbf{R}| \mathbf{T}_1, \mathbf{Z}_1)$ as the initial rotation distribution.


At each following frame, we first propagate the $N$ particles with the motion priors. Then the particles are updated with the latest observation $\mathbf{Z}_k$. Specifically, for each particle, the translation estimation $\mathbf{T}_k^i$ is used to compute the RoI of the object in image $\mathbf{Z}_k$. The resulting RoI is passed through the auto-encoder to compute the corresponding code. For each particle, the rotation distribution is updated with:
\begin{equation}
    P(\mathbf{R}_k|\mathbf{T}_k^i, \mathbf{Z}_{1:k}) \propto P(\mathbf{R}_k|\mathbf{T}_k^i, \mathbf{Z}_{k}) P(\mathbf{R}_k|\mathbf{R}_{k-1}), 
\end{equation}
where $P(\mathbf{R}_k|\mathbf{T}_k^i, \mathbf{Z}_{k})$ is the rotation distribution defined in Eq.~\eqref{eq:similarity}, and $P(\mathbf{R}_k|\mathbf{R}_{k-1})$ is the motion prior. Finally, we compute the posterior of the translation $P(\mathbf{T}_k^i|\mathbf{Z}_{1:k})$ with
\begin{equation}
    P(\mathbf{T}_k^i|\mathbf{Z}_{1:k}) \propto \underset{\mathbf{R}_k}\sum P(\mathbf{Z}_k|\mathbf{T}_k^i, \mathbf{R}_k)P(\mathbf{R}_k|\mathbf{T}_{1:k-1}^i, \mathbf{Z}_{1:k-1}),
\end{equation}
and use it as the weight $w^i$ of this particle. The systematic resampling method \cite{douc2005comparison} is used to resample the particles according to the weights $w^{1:N}$. Our 6D object pose tracking framework is shown in Fig.~\ref{fig:framework}.

Some robotic tasks require the expectation of the 6D pose of the object from the particle filter for decision making. The expectation can be represented as $(\mathbf{T}^E_k, \mathbf{R}^E_k)$. The translation expectation can be computed simply by averaging the translation estimations $\mathbf{T}^{1:N}_k$ for all the $N$ particles due to the uni-modal nature of translation in the object tracking task. Computing the rotation expectation $\mathbf{R}^E_k$ is less obvious since the distribution $P(\mathbf{R}_k)$ might be multi-modal and simply performing weighted averaging over all the discrete rotations is not meaningful. To compute the rotation expectation, we first summarize the rotation distribution for all the particles by taking the maximum probability for every discrete rotation, resulting in rotation distribution $P(\mathbf{R}^E)_k$. The rotation expectation $\mathbf{R}^E_k$ is then computed by weighted averaging the discrete egocentric rotations within a neighborhood of the previous rotation expectation $\mathbf{R}^E_{k-1}$ using the quaternion averaging method proposed in \cite{markley2007quaternion}. 

Performing codebook matching with the estimated RoIs also provides a way to detect tracking failures. We can first find the maximum similarity score among all the particles. Then if maximal score is lower than a pre-defined threshold, we determine it is a tracking failure. Algorithm~\ref{dpf_alg} summarizes our Rao-Blackwellized particle filter for 6D object pose tracking.

\begin{algorithm}[t]
    \setstretch{1.35}
    \SetKwInOut{Input}{input}\SetKwInOut{Output}{output}
    \SetAlgoLined
    \Input{
        $\mathbf{\mathbf{Z}}_{k}$, $(\mathbf{T}_{k-1}^{1:N}, P(\mathbf{R})_{k-1}^{1:N})$
    }
    \Output{
        $(\mathbf{T}_{k}^{1:N}, P(\mathbf{R})_{k}^{1:N})$
    }
    \Begin{
        $\{w^i\}_{i=1}^N\leftarrow \emptyset$ \;

            $(\bar{\mathbf{T}}_{k}^{1:N}, P(\bar{{\mathbf{R}}})_{k}^{1:N}) \leftarrow Propagate(\mathbf{T}_{k-1}^{1:N}, P(\mathbf{R})_{k-1}^{1:N})$\;
            \For{$(\bar{\mathbf{T}}^i_k, P(\bar{\mathbf{R}})^i_k) \in (\bar{\mathbf{T}}^{1:N}_k, P(\bar{\mathbf{R}})^{1:N}_k)$}{
                $P(\bar{\mathbf{R}})^i_k \leftarrow Codebook\_Match(\mathbf{Z}_k,  \bar{\mathbf{T}}^i_k) * P(\bar{\mathbf{R}})^i_k$\;
                $w^i \leftarrow Evaluate(\mathbf{Z}_k, \bar{\mathbf{T}}^i_k, P(\bar{\mathbf{R}}^i_k))$;
                
            }
            $(\mathbf{T}_{k}^{1:N}, P(\mathbf{R})_{k}^{1:N})\leftarrow Resample(\bar{\mathbf{T}}_{k}^{1:N}, P(\bar{\mathbf{R}})_{k}^{1:N}, \{w^i\}_{i=1}^N)$\;
    }
    \caption{6D Object Pose Tracking with PoseRBPF}
    \label{dpf_alg}
\end{algorithm}


\subsection{RGB-D Extension of PoseRBPF}
\label{sec:depth}
PoseRBPF can be extended to use depth measurements for computing the observation likelihoods. With the RGB image $\mathbf{Z}_k^C$ and the additional depth measurements $\mathbf{Z}_k^D$, the observation likelihood in Eq.~\eqref{eq:ob_likelihood} can be rewritten as:
\begin{align}
    & P(\mathbf{Z}_k|\mathbf{T}_k, \mathbf{R}_k) =  P(\mathbf{Z}_k^{C}, \mathbf{Z}_k^{D}|\mathbf{T}_k, \mathbf{R}_k) \nonumber \\
    & \propto P(\mathbf{R}_k|\mathbf{T}_k, \mathbf{Z}_k^{C})P(\mathbf{Z}_k^{C}|\mathbf{T}_k)P(\mathbf{Z}_k^{D}|\mathbf{T}_k).
\end{align}
Note that the auto-encoder only uses the RGB image. Therefore, $P(\mathbf{R}_k|\mathbf{T}_k, \mathbf{Z}_k^{C}, \mathbf{Z}_k^{D}) = P(\mathbf{R}_k|\mathbf{T}_k, \mathbf{Z}_k^{C})$. To compute the likelihood with the depth image $P(\mathbf{Z}_k^{D}|\mathbf{T}_k^i)$ for a translation hypothesis $\mathbf{T}_k^i$, we first render the object with pose $(\mathbf{T}_k^i, \mathbf{R}_k^*)$, where $\mathbf{R}_k^* = \underset{\mathbf{R}_k}{\arg\max}P(\mathbf{R}_k|\mathbf{T}_k^i, \mathbf{Z}_k^C)$ from the color image. By comparing the rendered depth image $\hat{\mathbf{Z}}^{Di}_k$ with the depth measurements $\mathbf{Z}_k^{D}$, we first estimate the visibility mask $\hat{V}^i_k = \{\forall p, \hat{\mathbf{Z}}^{Di}_k(p) - \mathbf{Z}_k^{D}(p) < m\}$, where $p$ indicates a pixel in the image and $m$ is a small positive constant margin to account for sensor noises. Therefore, the rendered pixel $p$ with depth less than $\mathbf{Z}_k^{D}(p) + m$ is determined as visible. With the estimated visibility mask, the \emph{visible depth discrepancy} between the two depth maps is computed as:
\begin{equation}
    \Delta_{k}^i(\hat{\mathbf{Z}}^{Di}_k, \mathbf{Z}_k^{D}, \hat{V}^i_k, \tau) = \underset{p\in\hat{V}^i_k}{\text{avg}}\Big(\min\big(\frac{|\mathbf{Z}_k^{D}(p) - \hat{\mathbf{Z}}^{Di}_k(p)|}{\tau}, 1\big)\Big),
\end{equation} 
where $\tau$ is a pre-defined threshold for each object. For every particle, we compute its \emph{depth score} $s_d^i = v^i_k(1 - \Delta^i_k)$, where $v^i_k$ is the visibility ratio of the object, i.e., the number of visible pixels according to the visibility mask divided by the total number of pixels rendered. Finally, we compute $P(\mathbf{Z}_k^{D}|\mathbf{T}_k^i)$ as $\phi'(s_d^i)$, where $\phi'(\cdot)$ is a gaussian probability density function centered at the maximum depth score among all the particles.

%% file: 3_experiments.tex
\section{EXPERIMENTS}

\begin{table} \setlength{\tabcolsep}{10pt}
	\centering
	\caption{Effect of the number of particles on frame rate in tracking.}
	\label{table:runtime}
	\begin{tabular}{|l|c|c|c|c|}
		\hline Number of particles & 50  & 100 & 200 & 400 \\
        \hline Frame rate (RGB) & 20.3 & 11.5 & 6.1 & 3.1 \\
        \hline Frame rate (RGB-D) & 14.8 & 9.5 & 5.0 & 2.8 \\  \hline
	\end{tabular}
	\vspace{-4mm}
\end{table}
 
\begin{table*}[h] \setlength{\tabcolsep}{6pt}
    \centering
    \caption{Results on YCB Video Dataset}
    \label{tab:ycb} 
    
    \begin{adjustbox}{width=\textwidth}
        \begin{tabular}{|c|cc|cc|cc|cc|cc|cc|cc|cc|}
            \hline
                                     & \multicolumn{10}{c|}{RGB}                                                                                                                                                        & \multicolumn{6}{c|}{RGB-D}                                                                                \\ \hline
                                    & \multicolumn{2}{c|}{PoseCNN~\cite{xiang2017posecnn}}  & \multicolumn{2}{c|}{DOPE~\cite{tremblay2018corl:dope}} & \multicolumn{2}{c|}{\begin{tabular}[c]{@{}c@{}}PoseRBPF\\ 50 particles\end{tabular}} & \multicolumn{2}{c|}{\begin{tabular}[c]{@{}c@{}}PoseRBPF\\ 200 particles\end{tabular}} & \multicolumn{2}{c|}{\begin{tabular}[c]{@{}c@{}}PoseRBPF++\\ 200 particles\end{tabular}} & \multicolumn{2}{c|}{PoseCNN+ICP~\cite{xiang2017posecnn}} & \multicolumn{2}{c|}{DenseFusion~\cite{wang2019densefusion}} & \multicolumn{2}{c|}{\begin{tabular}[c]{@{}c@{}}PoseRBPF\\ 200 particles\end{tabular}} \\ \hline
            objects                  & ADD           & ADD-S         & ADD           & ADD-S         & ADD         & ADD-S                & ADD              & ADD-S            & ADD               & ADD-S             & ADD             & ADD-S          & ADD        & ADD-S               & ADD              & ADD-S            \\ \hline
            002\_master\_chef\_can   & 50.9          & 84.0          & -             & -             & 56.1        & 75.6                 & 58.0             & 77.1             & \textbf{63.3}     & \textbf{87.5}     & 69.0            & 95.8           & -          & \textbf{96.4}       & \textbf{90.5}    & 95.1             \\
            003\_cracker\_box        & 51.7          & 76.9          & 55.9          & 69.8          & 73.4        & 85.2                 & 76.8             & 87.0             & \textbf{77.8}     & \textbf{87.6}     & 80.7            & 91.8           & -          & \textbf{95.5}       & \textbf{88.2}    & 93.0             \\
            004\_sugar\_box          & 68.6          & 84.3          & 75.7          & 87.1          & 73.9        & 86.5                 & 75.9             & 87.6             & \textbf{79.6}     & \textbf{89.4}     & \textbf{97.2}   & \textbf{98.2}  & -          & 97.5                & 92.9             & 95.5             \\
            005\_tomato\_soup\_can   & 66.0          & 80.9          & \textbf{76.1} & \textbf{85.1} & 71.1        & 82.0                 & 74.9             & 84.5             & 73.0              & 83.6              & 81.6            & 94.5           & -          & \textbf{94.6}       & \textbf{90.0}    & 93.8             \\
            006\_mustard\_bottle     & 79.9          & 90.2          & 81.9          & 90.9          & 80.0        & 90.1                 & 82.5             & 91.0             & \textbf{84.7}     & \textbf{92.0}     & \textbf{97.0}   & \textbf{98.4}  & -          & 97.2                & 91.9             & 96.3             \\
            007\_tuna\_fish\_can     & \textbf{70.4} & \textbf{87.9} & -             & -             & 56.1        & 73.8                 & 59.0             & 79.0             & 64.2              & 82.7              & 83.1            & 97.1           & -          & 96.6                & \textbf{91.1}    & 95.3             \\
            008\_pudding\_box        & \textbf{62.9} & \textbf{79.0} & -             & -             & 54.8        & 69.2                 & 57.2             & 72.1             & 64.5              & 77.2              & \textbf{96.6}   & \textbf{97.9}  & -          & 96.5                & 85.8             & 92.0             \\
            009\_gelatin\_box        & 75.2          & 87.1          & -             & -             & 83.1        & 89.7                 & \textbf{88.8}    & \textbf{93.1}    & 83.0              & 90.8              & \textbf{98.2}   & \textbf{98.8}  & -          & 98.1                & 96.3             & 97.5             \\
            010\_potted\_meat\_can   & \textbf{59.6} & \textbf{78.5} & 39.4          & 52.4          & 47.0        & 61.3                 & 49.3             & 62.0             & 51.8              & 66.9              & \textbf{83.8}   & \textbf{92.8}  & -          & 91.3                & 68.7             & 77.9             \\
            011\_banana              & \textbf{72.3} & \textbf{85.9} & -             & -             & 22.8        & 64.1                 & 24.8             & 61.5             & 18.4              & 66.9              & \textbf{91.6}   & \textbf{96.9}  & -          & 96.6                & 74.2             & 86.9             \\
            019\_pitcher\_base       & 52.5          & 76.8          & -             & -             & 74.0        & 87.5                 & \textbf{75.3}    & \textbf{88.4}    & 63.7              & 82.1              & \textbf{96.7}   & \textbf{97.8}  & -          & 97.1                & 86.8             & 94.2             \\
            021\_bleach\_cleanser    & 50.5          & 71.9          & -             & -             & 51.6        & 66.7                 & 54.5             & 69.3             & \textbf{60.5}     & \textbf{74.2}     & \textbf{92.3}   & \textbf{96.8}  & -          & 95.8                & 86.0             & 93.0             \\
            024\_bowl                & 6.5           & 69.7          & -             & -             & 26.4        & \textbf{88.2}        & \textbf{36.1}    & 86.0             & 28.4              & 85.6              & 17.5            & 78.3           & -          & 88.2                & \textbf{25.5}    & \textbf{94.2}    \\
            025\_mug                 & 57.7          & 78.0          & -             & -             & 67.3        & 83.7                 & 70.9             & 85.4             & \textbf{77.9}     & \textbf{89.0}     & 81.4            & 95.1           & -          & \textbf{97.1}       & \textbf{90.9}    & \textbf{97.1}    \\
            035\_power\_drill        & 55.1          & 72.8          & -             & -             & 64.4        & 80.6                 & 70.9             & \textbf{85.0}    & \textbf{71.8}     & 84.3              & 96.9            & 98.0           & -          & 96.0                & 93.9             & \textbf{96.1}    \\
            036\_wood\_block         & \textbf{31.8} & \textbf{65.8} & -             & -             & 0.0         & 0.0                  & 2.8              & 33.3             & 2.3               & 31.4              & \textbf{79.2}   & \textbf{90.5}  & -          & 89.7                & 20.1             & 89.1             \\
            037\_scissors            & 35.8          & 56.2          & -             & -             & 20.6        & 30.9                 & 21.7             & 33.0             & \textbf{38.7}     & \textbf{59.1}     & \textbf{78.4}   & 92.2           & -          & \textbf{95.2}       & 76.1             & 85.6             \\
            040\_large\_marker       & 58.0          & 71.4          & -             & -             & 45.7        & 54.1                 & 48.7             & 59.3             & \textbf{67.1}     & \textbf{76.4}     & 85.4            & 97.2           & -          & \textbf{97.5}       & \textbf{92.0}    & 97.1             \\
            051\_large\_clamp        & 25.0          & 49.9          & -             & -             & 27.0        & 73.2                 & \textbf{47.3}    & \textbf{76.9}    & 38.3              & 59.3              & \textbf{52.6}   & 75.4           & -          & 72.9                & 48.5             & \textbf{94.8}    \\
            052\_extra\_large\_clamp & 15.8          & 47.0          & -             & -             & 50.4        & 68.7                 & \textbf{26.5}    & \textbf{69.5}    & 32.3              & 44.3              & 28.7            & 65.3           & -          & 69.8                & \textbf{40.3}    & \textbf{90.1}    \\
            061\_foam\_brick         & 40.4          & 87.8          & -             & -             & 75.8        & 88.4                 & 78.2             & 89.7             & \textbf{84.1}     & \textbf{92.6}     & 48.3            & \textbf{97.1}  & -          & 92.5                & \textbf{81.1}    & 95.7             \\ \hline
            ALL                      & 53.7          & 75.9          & -             & -             & 57.1        & 74.8                 & 59.9             & 77.5             & \textbf{62.1}     & \textbf{78.4}     & 79.3            & 93.0           & -          & 93.1                & \textbf{80.8}    & \textbf{93.3}    \\ \hline
            \end{tabular}
    \end{adjustbox}
    \vspace{-2mm}
\end{table*}
 
\begin{table}[h] \setlength{\tabcolsep}{4pt}
    \centering
    \caption{T-Less Results: Object recall for $e_{vsd} < 0.3$ on all Primesense test scenes}
    \label{tab:tless} 
    \begin{adjustbox}{width=\textwidth/2}
        \begin{tabular}{|c|ccc|cc|cc|}
            \hline
            \multicolumn{6}{|c|}{Without GT 2D BBs}                                                                & \multicolumn{2}{c|}{\multirow{3}{*}{With GT 2D BBs}} \\ \cline{1-6}
            \multirow{3}{*}{Object} & \multicolumn{3}{c|}{RGB}                   & \multicolumn{2}{c|}{RGB-D}      & \multicolumn{2}{c|}{}                                \\ \cline{2-6}
                                    & SSD      & RetinaNet      & RetinaNet      & RetinaNet      & RetinaNet      & \multicolumn{2}{c|}{}                                \\ \cline{2-8} 
                                    & \cite{sundermeyer2018implicit} & \cite{sundermeyer2018implicit}       & PoseRBPF       & \cite{sundermeyer2018implicit} + ICP & PoseRBPF       & \cite{sundermeyer2018implicit}               & PoseRBPF                    \\ \hline
            1                       & 5.65     & 8.87           & \textbf{27.60} & 22.32          & \textbf{61.30} & 12.33                  & \textbf{80.90}              \\
            2                       & 5.46     & 13.22          & \textbf{26.60} & 29.49          & \textbf{63.10} & 11.23                  & \textbf{85.80}              \\
            3                       & 7.05     & 12.47          & \textbf{37.70} & 38.26          & \textbf{74.30} & 13.11                  & \textbf{85.60}              \\
            4                       & 4.61     & 6.56           & \textbf{23.90} & 23.07          & \textbf{64.50} & 12.71                  & \textbf{62.00}              \\
            5                       & 36.45    & 34.80          & \textbf{54.40} & 76.10          & \textbf{86.70} & 66.70                  & \textbf{89.80}              \\
            6                       & 23.15    & 20.24          & \textbf{73.00} & 67.64          & \textbf{71.50} & 52.30                  & \textbf{97.80}              \\
            7                       & 15.97    & 16.21          & \textbf{51.60} & 73.88          & \textbf{88.00} & 36.58                  & \textbf{91.20}              \\
            8                       & 10.86    & 19.74          & \textbf{37.90} & 67.02          & \textbf{84.00} & 22.05                  & \textbf{95.60}              \\
            9                       & 19.59    & 36.21          & \textbf{41.60} & 78.24          & \textbf{86.00} & 46.49                  & \textbf{77.10}              \\
            10                      & 10.47    & 11.55          & \textbf{41.50} & \textbf{77.65} & 74.30          & 14.31                  & \textbf{85.30}              \\
            11                      & 4.35     & 6.31           & \textbf{38.30} & 35.89          & \textbf{62.60} & 15.01                  & \textbf{89.50}              \\
            12                      & 7.80     & 8.15           & \textbf{39.60} & 49.30          & \textbf{71.00} & 31.34                  & \textbf{91.20}              \\
            13                      & 3.30     & 4.91           & \textbf{20.40} & 42.50          & \textbf{42.10} & 13.60                  & \textbf{89.30}              \\
            14                      & 2.85     & 4.61           & \textbf{32.00} & 30.53          & \textbf{50.10} & 45.32                  & \textbf{70.20}              \\
            15                      & 7.90     & 26.71          & \textbf{41.60} & \textbf{83.73} & 76.60          & 50.00                  & \textbf{96.60}              \\
            16                      & 13.06    & 21.73          & \textbf{39.10} & 67.42          & \textbf{83.80} & 36.09                  & \textbf{97.00}              \\
            17                      & 41.70    & \textbf{64.84} & 40.00          & \textbf{86.17} & 78.40          & 81.11                  & \textbf{87.00}              \\
            18                      & 47.17    & 14.30          & \textbf{47.90} & \textbf{84.34} & 81.10          & 52.62                  & \textbf{89.70}              \\
            19                      & 15.95    & 22.46          & \textbf{40.60} & 50.54          & \textbf{61.80} & 50.75                  & \textbf{83.20}              \\
            20                      & 2.17     & 5.27           & \textbf{29.60} & 14.75          & \textbf{55.00} & 37.75                  & \textbf{70.00}              \\
            21                      & 19.77    & 17.93          & \textbf{47.20} & 40.31          & \textbf{72.70} & 50.89                  & \textbf{84.40}              \\
            22                      & 11.01    & 18.63          & \textbf{36.60} & 35.23          & \textbf{63.80} & 47.60                  & \textbf{77.70}              \\
            23                      & 7.98     & 18.63          & \textbf{42.00} & 42.52          & \textbf{82.40} & 35.18                  & \textbf{85.90}              \\
            24                      & 4.74     & 4.23           & \textbf{48.20} & 59.54          & \textbf{83.20} & 11.24                  & \textbf{91.80}              \\
            25                      & 21.91    & 18.76          & \textbf{39.50} & 70.89          & \textbf{77.70} & 37.12                  & \textbf{88.70}              \\
            26                      & 10.04    & 12.62          & \textbf{47.80} & 66.20          & \textbf{85.00} & 28.33                  & \textbf{90.90}              \\
            27                      & 7.42     & 21.13          & \textbf{41.30} & \textbf{73.51} & 68.00          & 21.86                  & \textbf{79.10}              \\
            28                      & 21.78    & 23.07          & \textbf{49.50} & 61.20          & \textbf{79.30} & 42.58                  & \textbf{72.10}              \\
            29                      & 15.33    & 26.65          & \textbf{60.50} & 73.04          & \textbf{86.30} & 57.01                  & \textbf{96.00}              \\
            30                      & 34.63    & 29.58          & \textbf{52.70} & \textbf{92.90} & 80.10          & 70.42                  & \textbf{77.00}              \\ \hline
            Mean                    & 14.67    & 18.35          & \textbf{41.67} & 57.14          & \textbf{73.16} & 36.79                  & \textbf{85.28}              \\ \hline
            \end{tabular}
    \end{adjustbox}
        \vspace{-4mm}
    \end{table}

\subsection{Datasets}
We evaluated our method on two datasets: the YCB Video dataset \cite{xiang2017posecnn} and the T-LESS dataset \cite{hodan2017t}.

\vspace{2mm}

\noindent {\bf YCB Video dataset:} The YCB video dataset contains RGB-D video sequences of 21 objects from the YCB Object and Model Set \cite{Calli-2015-5995}. It contains textured and textureless household objects put in different arrangements. Objects are annotated with 6D object poses and two metrics are used for quantitative evaluation. The first metric is ADD, which is the average distance between the corresponding 3D points on the object at groundtruth pose vs the predicted pose. The second metric is ADD-S, which is the average distance between the \emph{closest point} between the 3D model of the object at groundtruth and the model of the object at the predicted pose.   ADD-S is designed for symmetric objects, since it focuses on shape matching, rather than exact pose matching. 


\vspace{2mm}

\noindent {\bf T-LESS:} This dataset contains RGB-D sequences of 30 non-textured industrial objects. Evaluation is done on 20 test scenes. The dataset is challenging because the objects do not have texture and they have various forms of symmetries and occlusions. We follow the evaluation pipeline in SIXD challenge and used Visible Surface Discrepancy $err_{vsd}$ \cite{VSD_ECCV16} to evaluate the quality of the pose estimation. Visual surface discrepancy is computed as mean average of the distance between the visible points. The metric is the recall of correct 6D poses where $err_{vsd} < 0.3$ with tolerance $20mm$ and visibility of more than $10\%$.

\subsection{Implementation Details}
The auto-encoder is trained for each object separately for $150,000$ iterations with batch size of $64$ using the Adam optimizer with learning rate of $0.0002$. The auto-encoder is optimized with the L2 loss on the $N$ pixels with largest reconstruction errors. Larger $N$s are more suitable for textured objects to capture more details. We use $N=2000$ for textured objects and $N=1000$ for non-textured objects. The training data is generated by rendering the object at random rotation and superimposed at random crops of the MS-COCO dataset \cite{lin2014microsoft} at resolution $128\times128$. In addition to the target object, three additional objects are sampled at random locations and scales to generate training data with occlusions. The target object is positioned at the center of the image and jittered with 5 pixels, the object is sampled uniformly at scales between $0.975$ and $1.025$ with random lighting. Color is randomized in HSV space and we also add Gaussian noise to pixel values to reduce the gap between the real and synthetic data. The images are rendered online for each training step to provide a more diverse set of training data. The architecture of the network is described in \cite{sundermeyer2018implicit}. It consists of four $5 \times 5$ convolutional layers and four $5 \times 5$ deconvolutional layers for the encoder and the decoder, respectively. The standard deviations used to compute observation likelihoods in Eq.~\eqref{eq:similarity} are selected between 0.03 and 0.1. The codebook for each object is pre-computed offline and loaded during test time. Computation of observation likelihood is done efficiently on a GPU. Table \ref{table:runtime} shows the frame rate at which \prbpf\ can process images.

\begin{figure}
    \begin{tabular}{c}
         \includegraphics[width=0.4\textwidth]{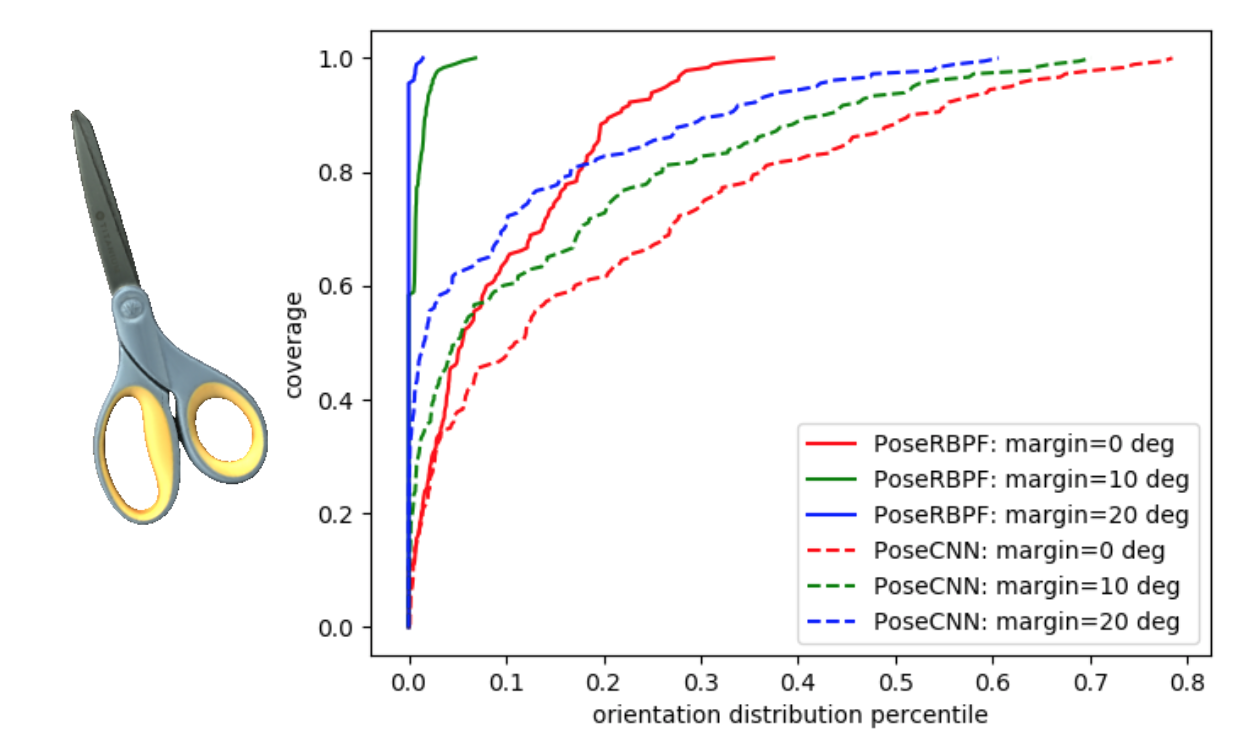}  \\
         \includegraphics[width=0.4\textwidth]{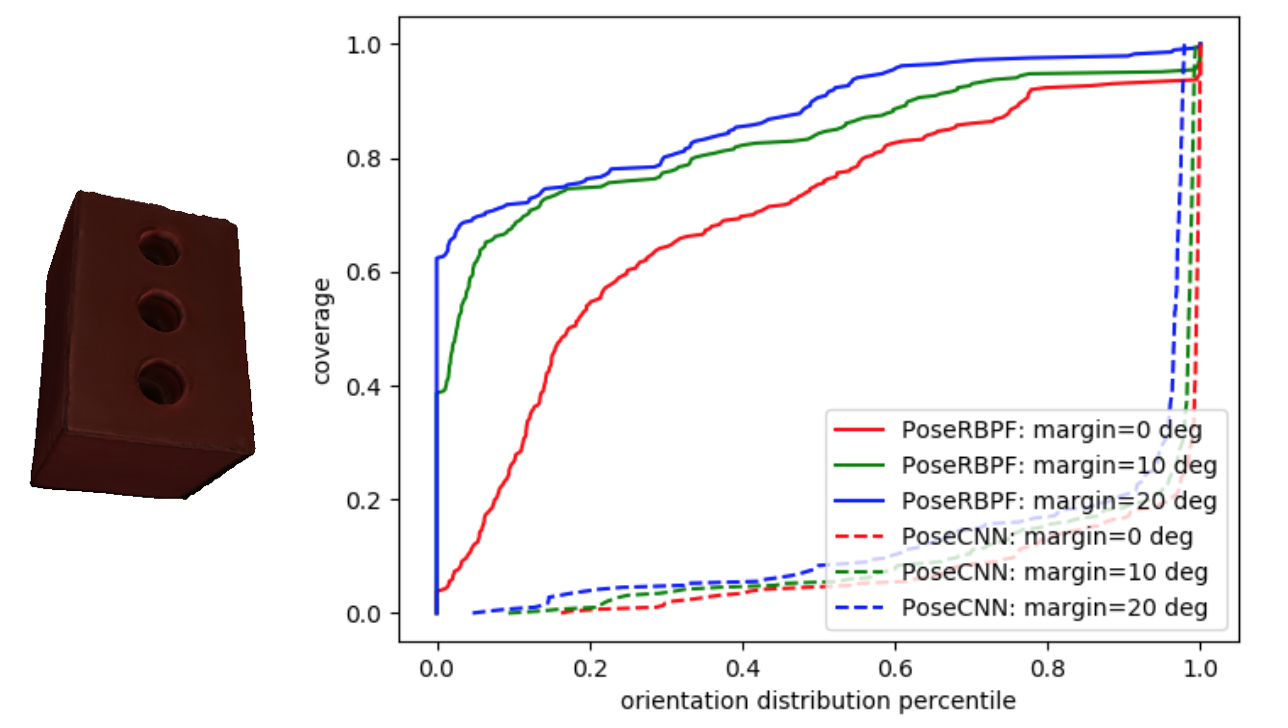}
    \end{tabular}
    \caption{Rotation Coverage Percentile comparison between PoseRBPF and PoseCNN for scissors and foam brick. Foam brick has $180^\circ$ planar rotation and scissors is an  asymmetric object.}
    \label{fig:distribution}
    \vspace{-4mm}
\end{figure}

\subsection{Results on YCB Video Dataset}
Table \ref{tab:ycb} shows the pose estimation results on the YCB video dataset, where we compare with the state-of-the-art methods for pose estimation using RGB images~\cite{xiang2017posecnn, tremblay2018corl:dope} and RGB-D images~\cite{xiang2017posecnn, wang2019densefusion}.
We initialize \prbpf\ using PoseCNN at the first frame or after the object was heavily occluded. On average, this happened only  $1.03$ times per sequence. As can be seen, our method significantly improves the accuracy of 6D pose estimation when using 200 particles. Note that our method handles symmetric objects such as 024\_bowl, 061\_foam\_brick much better. One of the objects on which \prbpf\ performs poorly is 036\_wood\_block, which is caused by the difference in texture of the 3D model of the wooden block and the texture of the wooden block used in the real images. In addition, the physical dimensions of the wooden block are different between real images and the model contained in this dataset. Another observation is that with the increase in the number of particles, the accuracy improves significantly because with more samples the variations in scale and translation of an object are covered much better. 
\begin{figure*}
    \centering
    \begin{tabular}{cc}
         \includegraphics[width=0.40\textwidth]{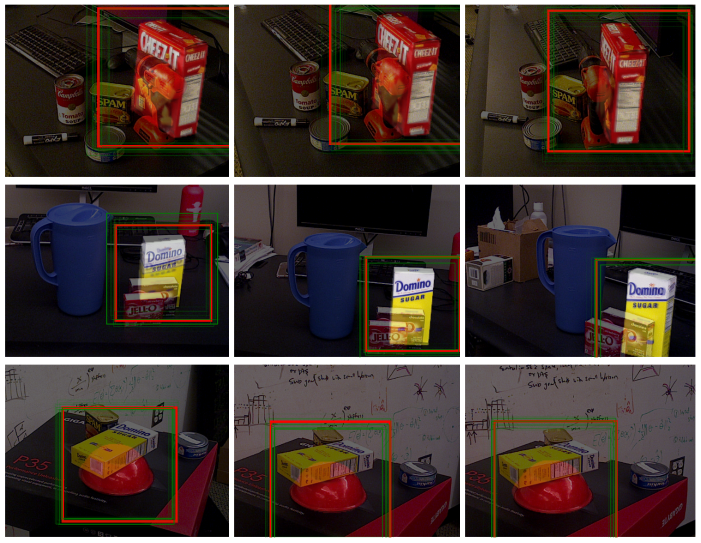} & 
         \includegraphics[width=0.40\textwidth]{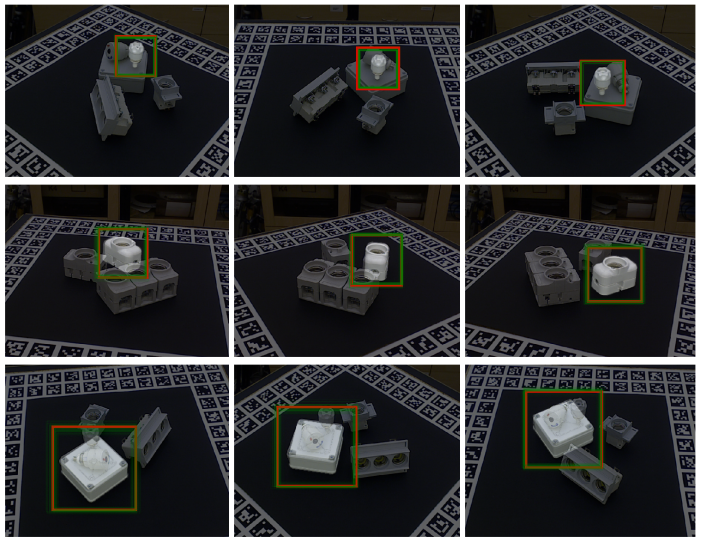}
    \end{tabular}
    \caption{Visualization of estimated poses on the YCB Video dataset (left) and T-LESS dataset (right). Ground truth bounding boxes are red, green bounding boxes are particle RoIs, and the object models are superimposed on the images at the pose estimated by \prbpf. }
    \label{fig:qualitatitve}
    \vspace{-6mm}
\end{figure*}

It has been shown in the context of robot localization that adding samples drawn according to the most recent  observation can improve the localization performance~\cite{PRbook}.  Here, we applied such a technique by  sampling $50\%$ of the particles around PoseCNN predictions and  the other $50\%$ from the particles of the previous time step. Our results show that such a hybrid version, PoseRBPF++, further improves the pose estimation accuracy of our approach.
Fig.~\ref{fig:qualitatitve} illustrates the 6D pose estimation on YCB Video dataset. Depth measurements contain useful information to improve the pose estimation accuracy. By comparing the depth of the object rendered at the estimated pose and the depth image (explained in Sec \ref{sec:depth}), our method achieves the state-of-the-art performance. The comparison between RGB and RGB-D versions of PoseRPBF shows using depth information with the same number of particles improves the accuracy of estimated poses significantly. Note that depth information is only used during inference and the encoder takes only the RGB images.


\vspace{-2mm}
\subsection{Results on T-LESS Dataset}
\begin{figure*}
    \centering
    \begin{tabular}{cc}
         \includegraphics[width=0.35\textwidth]{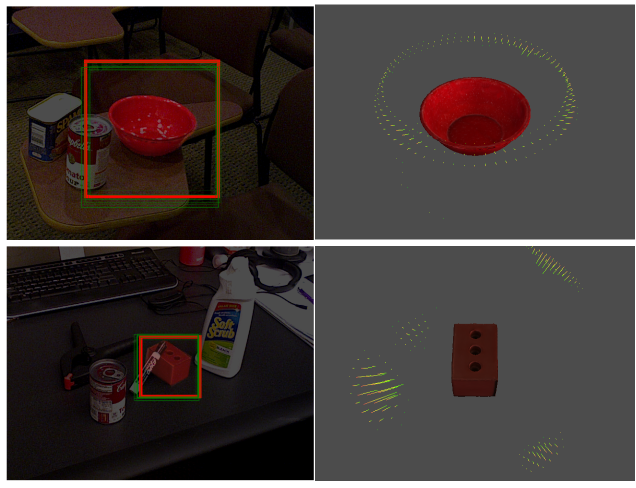} &
         \includegraphics[width=0.35\textwidth]{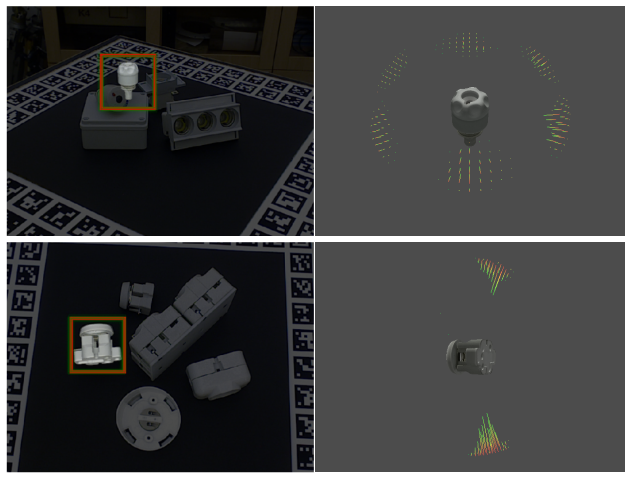}
    \end{tabular}
    \caption{Visualization of rotation distributions. For each image, the distribution over the rotation is visualized. The lines represent the probability for rotations that are higher than a threshold. The length of each line is proportional to the probability of that viewpoint.  As can be seen, \prbpf\ naturally represents uncertainties due to various kinds of symmetries, including rotational symmetry of the bowl, mirror symmetry of the foam brick, and discrete rotational symmetries of the T-LESS objects on the right.}
    \label{fig:vis_distribution}
    \vspace{-4mm}
\end{figure*}

Table \ref{tab:tless} presents our results on the T-LESS dataset. T-LESS is a challenging dataset because objects do not have texture and objects are occluded frequently in different frames. We compared our method with \cite{sundermeyer2018implicit} which uses a similar auto-encoder, but does not use any temporal information. We evaluated both using ground truth bounding boxes and the detection output from RetinaNet \cite{lin2018focal} that is used in \cite{sundermeyer2018implicit}. Our tracker uses 100 particles, and is reinitialized whenever the observation likelihood drops below a threshold. The results show that the recall for correct object poses doubles by tracking the object pose rather than just predicting object pose from single images in the RGB case. With additional depth images, the recall can be further improved by around 76\%, and PoseRBPF outperforms refining \cite{sundermeyer2018implicit} with ICP by 28\%. For the experiments with ground truth bounding boxes, rotation is tracked using the particle filter and translation is inferred from the scale of the ground truth bounding box. This experiment highlights the viewpoint accuracy. In this setting, recall increases significantly for all the methods and the particle filter consistently outperforms \cite{sundermeyer2018implicit}, which shows the importance of temporal tracking for object pose estimation. Fig.~\ref{fig:qualitatitve} shows the 6D pose estimation of PoseRBPF on several T-LESS images.

\subsection{Analysis of Rotation Distribution}
Unlike other 6D pose estimation methods that output a single estimate for the 3D rotation of an object, PoseRBPF tracks full distributions over object rotations. Fig.~\ref{fig:vis_distribution} shows  example distributions for some objects. There are two types of uncertainties in these distributions. The first source is the symmetry of the objects resulting in multiple poses with similar appearances. As expected, each cluster of the viewpoints corresponds to one of the similarity modes. The variance for each cluster corresponds to the true uncertainty of the pose. For example for the bowl, each ring of rotations corresponds to the uncertainty around the azimuth because the bowl is a rotationally symmetric object. Different rings show the uncertainty on the elevation. 

To measure how well PoseRBPF's capture rotation uncertainty, we compared  PoseRBPF estimates to those of PoseCNN assuming a Gaussian uncertainty with mean at the PoseCNN estimate. 
Fig.~\ref{fig:distribution} shows this comparison for the scissors and foam brick objects.  Here, the x-axis ranges over percentiles of the rotation distributions, and the y-axis shows how often the ground truth pose is within 0, 10, or 20 degrees of one of the rotations contained in the corresponding percentile.  For instance, for the scissors, the red, solid line indicates that 80\% of the time, the ground truth rotation is within 20 degrees of an rotation taken from the top 20\% of the PoseRBPF distribution. If we take the top 20\% rotations estimated by PoseCNN assuming a Gaussian uncertainty, this number drops to about 60\%, as indicated by the lower dashed, red line.  The importance of maintaining multi-modal uncertainties becomes even more prominent for the foam brick, which has a $180^\circ$ symmetry.  Here, PoseRBPF achieves high coverage, whereas PoseCNN fails to generate good rotation estimates even when moving further from the generated estimate.